\RequirePackage{amsmath}
\documentclass[runningheads]{llncs}

\usepackage[utf8]{inputenc} 
\usepackage[T1]{fontenc}    
\usepackage{hyperref}       
\usepackage{url}            
\usepackage{booktabs}       
\usepackage{amsfonts}       
\usepackage{nicefrac}       
\usepackage{microtype}      
\usepackage{lipsum}
\usepackage{fancyhdr}       
\usepackage{graphicx}       
\usepackage{multirow}
\usepackage{comment}
\usepackage{orcidlink}

\usepackage{amssymb}
\usepackage{pifont}
\newcommand{\cmark}{\ding{51}}%
\newcommand{\xmark}{\ding{55}}%
\usepackage[T1]{fontenc}
\usepackage{amsmath}
\usepackage{xcolor}

\definecolor{c1}{HTML}{1034A6}
\definecolor{c2}{HTML}{412F88}
\definecolor{c3}{HTML}{722B6A}
\definecolor{c4}{HTML}{A2264B}
\definecolor{c5}{HTML}{D3212D}
\definecolor{c6}{HTML}{F62D2D}
\usepackage{graphicx}
\begin{document}
\title{Pitfalls of topology-aware image segmentation}
%
%
\author{
Alexander H. Berger\thanks{Equal Contribution}\inst{1, 2}\orcidlink{0009-0004-8843-7684} \and
Laurin Lux*\inst{1,2,3}\orcidlink{0009-0003-7359-6212} \and
Alexander Weers \inst{1,2}\and
Martin Menten \inst{1,2,3,4} \orcidlink{0000-0001-8261-7810}\and 
Daniel Rueckert\inst{1, 2, 3, 4}\orcidlink{0000-0002-5683-5889} \and
Johannes C. Paetzold\inst{4,5}\orcidlink{0000-0002-4844-6955}}
%
\authorrunning{Berger \& Lux et al.}
%
\institute{
Chair for AI in Healthcare and Medicine, Technical University of Munich (TUM) and TUM University Hospital, Germany \and 
School of Computation, Information and Technology, TUM, Germany \and
Munich Center for Machine Learning (MCML), Munich, Germany \and
Department of Computing, Imperial College London, UK \and
Weill Cornell Medicine, Cornell University, New York City, USA
}
\maketitle              
\begin{abstract}

Topological correctness, i.e., the preservation of structural integrity and specific characteristics of shape, is a fundamental requirement for medical imaging tasks, such as neuron or vessel segmentation.
Despite the recent surge in topology-aware methods addressing this challenge, their real-world applicability is hindered by flawed benchmarking practices. In this paper, we identify critical pitfalls in model evaluation that include inadequate connectivity choices, overlooked topological artifacts in ground truth annotations, and inappropriate use of evaluation metrics. Through detailed empirical analysis, we uncover these issues' profound impact on the evaluation and ranking of segmentation methods. Drawing from our findings, we propose a set of actionable recommendations to establish fair and robust evaluation standards for topology-aware medical image segmentation methods. \footnote[1]{Code is available at \url{https://github.com/AlexanderHBerger/topo-pitfalls}}

\end{abstract}
\section{Introduction}
Quantitative imaging biomarkers are of increasing importance in modern medicine. The development and evaluation of these biomarkers are directly linked to the emerging capabilities of artificial intelligence (AI) models \cite{panayides2020ai}. Rapid progress has been made in medical image segmentation with architectures such as the Unet \cite{ronneberger2015u} or the Vision Transformer \cite{dosovitskiy2020image}. Despite these advances, achieving perfect pixel-wise accuracy often remains impossible in image segmentation tasks. Consequently, many studies investigate the quality of these segmentations beyond purely pixel-based performance metrics \cite{bohlender2021survey}. 

\begin{figure}[t]
    \centering
    \includegraphics[width=0.85\linewidth, trim={0 0 0 0.45cm},clip]{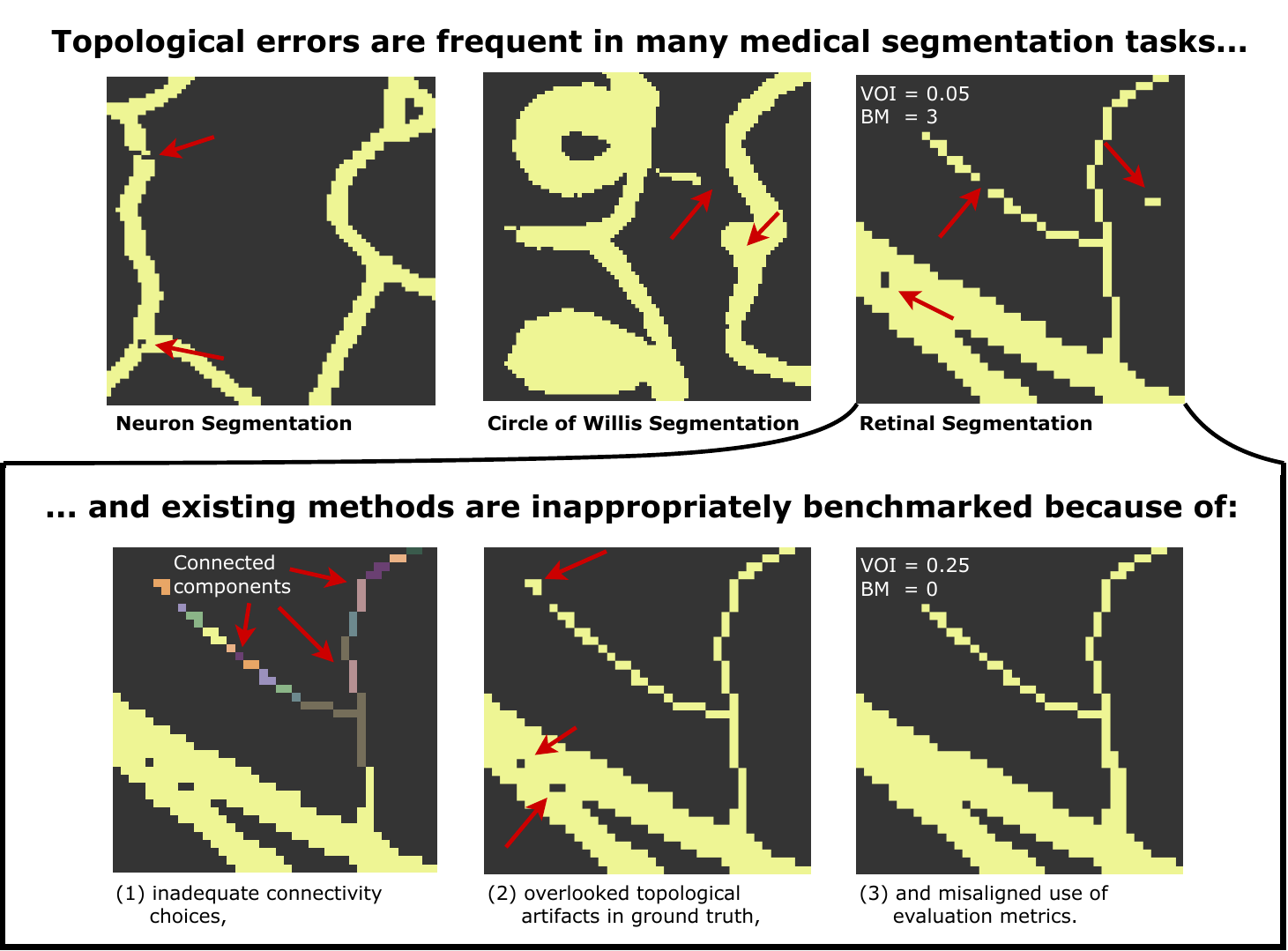}
    \vspace{-0.4cm}
    \caption{Topological errors are present across distinct medical image segmentation tasks, e.g., in neuron, Circle of Willis, and retinal segmentation (top). We identify three critical pitfalls (bottom) in the evaluation of topology-aware segmentation methods. These include inadequate connectivity choices that misrepresent a dataset's semantics (left, e.g., representing a single vessel as multiple components), overlooked topological artifacts that skew evaluation results (center), and misaligned use of evaluation metrics that lack expressive power (right, e.g., VOI entangles volumetric and topological information).}
    \vspace{-0.5cm}
    \label{fig:overview}
\end{figure}

An important property in image segmentation is topological correctness, which considers structural integrity aside from volumetric accuracy. Its significance has been identified separately in several areas of medical imaging, such as radiology, angiology, and neurology. For example, in lesion detection, metrics such as \textit{lesion true positive rate} and \textit{lesion false positive rate} are reported on an instance level, which are closely related to the topological features of a segmentation. \cite{carass2017longitudinal,yan2018deeplesion,zhang2021geometric}. In ophthalmology, the \textit{connectivity} metric was proposed by Gegundez et al. \cite{gegundez2011function} to measure the essential connectivity information in a predicted vessel segmentation. In neuron segmentation, community efforts led to the discovery that two topology-related metrics, \textit{adapted rand error (ARE)} and \textit{variation of information (VOI)}, are closely tied to expert assessment of the quality of predicted neuron segmentations \cite{arganda2015crowdsourcing}. Moreover, \textit{split/merge errors} are used as essential quality estimates for neuron segmentation \cite{januszewski2018high}.

The importance of topological correctness in all these medical fields has led to numerous methods that focus on preserving the topology of a target structure in image segmentation. These methods aim to find general solutions for topologically accurate image segmentation and have achieved impressive results in diverse medical tasks \cite{hu2019topology,hu2021topology,hu2022structure,mosinska2018beyond,stucki2023topologically,li2023robust,berger2024topologically,lux2024topograph}. However, we uncover that many works overlook the particularities of topological evaluation and the corresponding downstream tasks when comparing different methods. In particular, we demonstrate that the benchmarking of topology-aware segmentation methods is negatively affected by (1) \textit{connectivity choices that are inadequate for the underlying data,} (2) \textit{overlooked topological artifacts in the ground truth labels that can bias results,} and (3) \textit{inappropriate use of evaluation metrics}. These practices result in the reporting of inaccurate absolute and relative performance of different methods and hinder a fair and robust assessment of their suitability to a specific task or domain. This work provides a detailed overview of these issues and analyses the impact of the most common pitfalls in an extensive empirical study. Based on our findings, we propose several solutions to these issues.

\section[Related Work]{Related Work}
\subsubsection{Topology-aware image segmentation methods}
Numerous studies have tried to enhance segmentations' topological integrity in highly task-specific settings. Our study focuses on works that propose general-purpose solutions for improving topological correctness in multiple domains. Many of these works build on persistent homology (PH) and its differentiability to define loss functions for neural network optimization \cite{hu2019topology,stucki2023topologically,byrne2022persistent,clough2020topological,berger2024topologically}. These methods are based on the computation of persistence diagrams using filtered cubical complexes from either T- or V-construction. In the persistence diagram, the birth and death cells can be identified and used to calculate a loss. Notably, for all these methods, the choice of cubical complex construction strongly impacts the persistence diagrams and, thus, the loss \cite{bleile2022persistent}. Other methods use discrete Morse theory (DMT) to achieve topology-aware image segmentation \cite{banerjee2020semantic,hu2022structure}. Apart from PH and DMT methods, other methods, including delineation \cite{mosinska2018beyond}, post-processing \cite{li2023robust,li2024universal}, homotopy warping \cite{hu2022structure}, skeletonization \cite{shit2021cldice}, and component graphs \cite{lux2024topograph} were proposed.

\subsubsection{Topological evaluation metrics}
Most studies on topologically accurate segmentation use a combination of different metrics to showcase their method's effectiveness. These metrics usually consist of pixel-wise metrics and topological metrics. A pixel-wise metric (e.g., cross-entropy or Dice coefficient) measures pixel-wise agreement between the prediction and label, disregarding topological characteristics. Although a perfect pixel-wise agreement implies identical topology, these metrics are not interpretable for values $\neq 1$ from a topological perspective. 

Several topological metrics have been specifically proposed for the task of neuron segmentation. They mainly revolve around measuring the number of \textbf{split/merge errors}, which are concepts related to topological errors in dimension 0. Two widespread metrics are Rand Index (RI)-based metrics (e.g., ARE or adjusted rand index (ARI)) \cite{hubert1985comparing,arbelaez2010contour,unnikrishnan2007toward}, and the VOI metric \cite{meilua2003comparing,meilǎ2005comparing,meilua2007comparing}, both of which were originally proposed to measure cluster similarity. They have been adapted to image segmentation by partitioning a likelihood map to an instance map, where all pixels belonging to the same connected component are viewed as one cluster. Then, RI-related metrics measure pair-wise pixel agreement in the ground truth and prediction \cite{arbelaez2010contour,unnikrishnan2007toward}. VOI measures how much information about a pixel's instance in the prediction can be gained by the pixel's instance in the ground truth and vice versa \cite{nunez2013machine}.

Lastly, purely topological metrics have been proposed for measuring topological accuracy. These metrics include the Betti number error \cite{hu2019topology}, the Betti matching error \cite{stucki2023topologically}, and the DIU metric \cite{lux2024topograph}. These metrics transfer the images to topological spaces and measure the number of topological errors. They give an interpretable quantification of the topological accuracy and provide various degrees of rigor: while the Betti number error captures global agreement between the number of topological features in each dimension \cite{hu2019topology}, the Betti matching error takes spatial correspondence of these features into account \cite{stucki2023topologically}. The DIU metric additionally measures the correspondence between the topological features in union and the intersection of an image pair \cite{lux2024topograph}. Although these metrics capture interpretable topological information, they disregard volumetric information of topological features, which is unsuitable for certain downstream tasks.

\subsubsection{Commonly used benchmarking datasets}

The importance of different topological characteristics is deeply tied to the semantics of a specific domain and the associated downstream tasks. Some datasets frequently appear in method papers on topology-aware segmentation. In medical imaging, neuron and vessel segmentation datasets are commonly used as benchmarks, whereas in computer vision, aerial imaging datasets are often used. The most prominent datasets are:

\noindent{\textit{DRIVE}.} The DRIVE dataset \cite{DRIVE_dataset}, used in \cite{hu2022structure,hu2019topology,hu2021topology,shit2021cldice,attari2023multi,wu2024deep}, consists of color fundus images of the human retina. The blood vessels are represented as the foreground (FG) components. The background (BG) components can be interpreted as inter-vessel areas. However, it is important to note that the loops around the background components are often a result of the 2D projection of the vasculature and commonly do not encode physiologically connected vessels. 

\noindent{\textit{CREMI}.} The CREMI dataset \cite{funke2018large}, used in \cite{li2023robust,hu2019topology,hu2021topology,stucki2023topologically,shit2021cldice,hu2022structure,lux2024topograph}, contains brain images visualized using transmission electron microscopy. Boundary maps are commonly used as data representations to evaluate topology-aware methods. Here, foreground components resemble neuron boundaries, while the background components resemble two structures: foremost neurons, and a smaller fraction of the background components resemble synaptic clefts. Some works also use the inverse map, where the foreground resembles neurons and synaptic clefts.

\noindent{\textit{Roads}.} The Roads dataset \cite{roads_dataset}, used in \cite{mosinska2018beyond,hu2019topology,hu2021topology,stucki2023topologically,shit2021cldice,lux2024topograph}, contains RGB satellite images where the ground truth represents road networks. It has also become a popular benchmark dataset in medical image segmentation due to its complex topological properties. In the Roads dataset, preserving connectivity—captured by topological features in dimension 1—is crucial for ensuring access between different areas of the map.

Topological information is essential for many other datasets that are not commonly used for benchmarking. We provide a non-exhaustive list of such datasets below. Examples for lesion detection are the MSSEG2 \cite{commowick2021msseg} and ISBI2015 \cite{carass2017longitudinal} challenge datasets. In neuron segmentation, the SNEMI3D \cite{kasthuri2015saturated} and ISBI12 \cite{arganda2015crowdsourcing} are other popular datasets. In ophthalmology, the FIVES \cite{jin2022fives}, STARE \cite{hoover2000locating}, and ROSE \cite{ma2020rose} datasets are common datasets to benchmark segmentation performance on color fundus and optical coherence tomography angiography images. The TopCow \cite{yang2023topcow} and VesSAP \cite{todorov2020machine} datasets are frequently used human and murine brain vessel datasets. In this work, we focus on the three most commonly used datasets (CREMI, DRIVE, Roads) as examples to investigate the impact of the identified pitfalls: \textit{ (1) Wrong Connectivity Choice}, \textit{(2) Ground Truth Artifacts}, and \textit{(3) Wrong Evaluation Metrics}.

\begin{table}[b]
    \scriptsize
    \centering
    \renewcommand{\arraystretch}{1.1}
    \setlength{\tabcolsep}{4pt}
    \caption{Dependence of the number of connected components on the foreground connectivity choice. The background connectivity is set to the opposite connectivity choice to satisfy the Jordan closed curve theorem. \textbf{A} denotes that all connectivity is selected for the foreground and direct for the background. \textbf{D} is used to denote the inverted connectivity setting.}
    \label{tab:CC_tab}
    \begin{tabular}{lcccccccccc}
    \toprule
     & \multicolumn{2}{c}{\textbf{DRIVE-2D}} & \multicolumn{2}{c}{\textbf{CREMI-2D}} & \multicolumn{2}{c}{\textbf{Roads-2D}} & \multicolumn{2}{c}{\textbf{MSSEG2-3D}} \\
    \cmidrule(lr){2-3} \cmidrule(lr){4-5} \cmidrule(lr){6-7} \cmidrule(lr){8-9}
     & \textbf{FG} & \textbf{BG} & \textbf{FG} & \textbf{BG} & \textbf{FG} & \textbf{BG} & \textbf{FG} & \textbf{BG} \\
    \midrule
    \textbf{A}     & 132   & 2362   & 705    & 49592   &  641   &  8839 & 153  & 29   \\
    \textbf{D}     & 18850 & 1113   & 712    & 44815   &  644   &  7535 & 155  & 29   \\
    \textbf{Ratio} & 0.8\% & 47.1\% & 99.0\% & 90.4\% & 99.5\% & 85.2\% & 98.7 & 100  \\
    \bottomrule
    \vspace{-\intextsep}
    \end{tabular}
    \end{table}

\section{Common Pitfalls}
\subsection{\textit{Connectivity choices} distort the performance ranking between different methods}
\label{sec:connectivity}
\begin{figure}[t]
    \centering
    \includegraphics[width=1\linewidth, trim={0 0.7cm 0 0.5cm},clip]{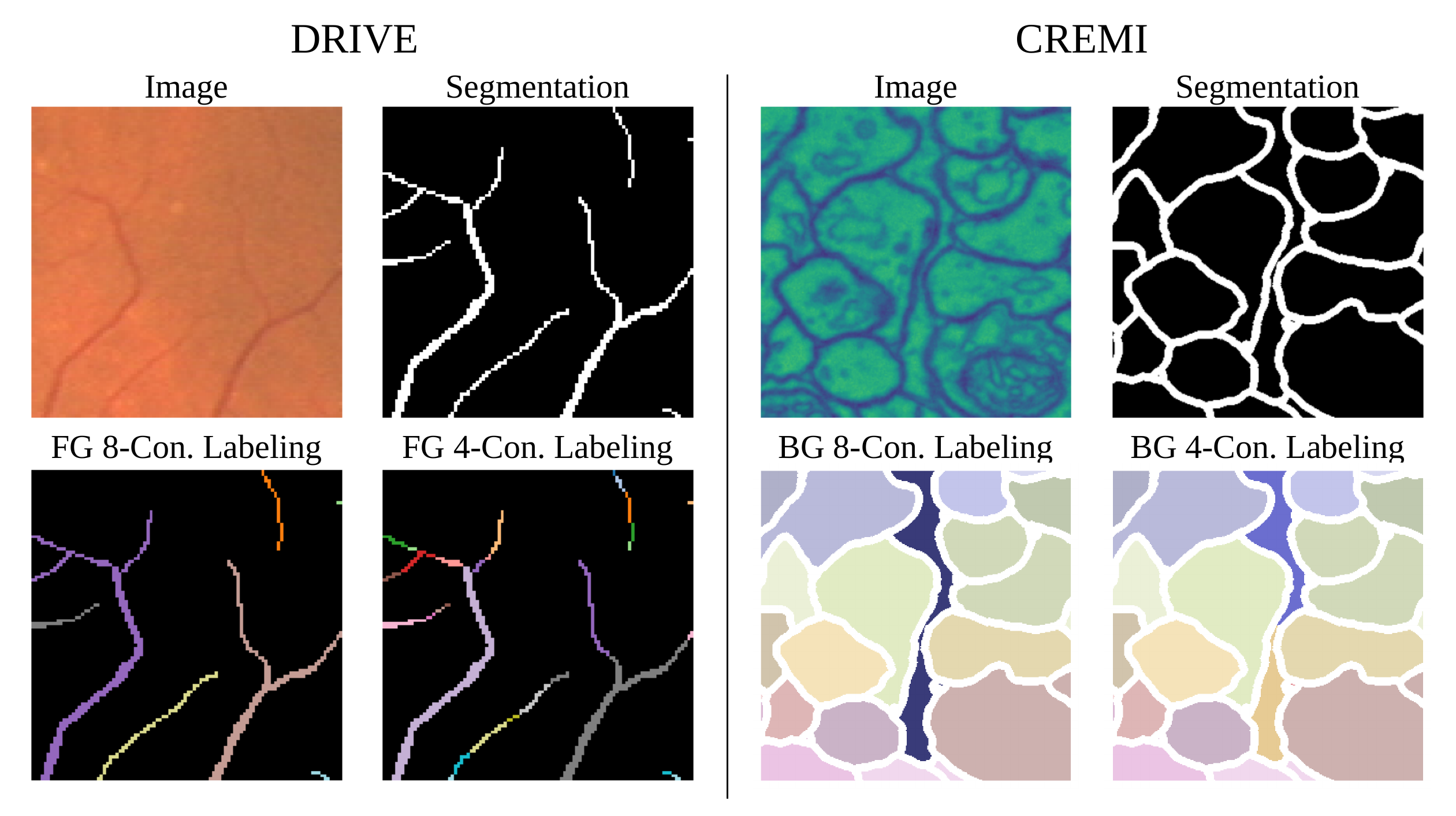}
    \caption{Example of the importance of making the correct connectivity choices for the DRIVE and CREMI datasets. In the DRIVE dataset, small vessels are disconnected with 4-connectivity for the FG. In the CREMI dataset, synaptic clefts can become disconnected with 4-connectivity for the BG.}
    \label{fig:2_1_connectivity}
\end{figure}

Voxel connectivity strongly affects topological representations and, hence, directly impacts model training and evaluation. Most prior works do not report which connectivity they use or appear to use an unfavorable choice where topological features lose their semantic meaning (see Fig. \ref{fig:2_1_connectivity}).

To translate a discrete 2D or 3D binary image $\mathcal{I}$ into a topological space, defining the connectivity between voxels (i.e., whether diagonal adjacent voxels belong to the same component) is necessary. Voxel connectivity in a $D$-dimensional image can be defined by {\em direct} connectivity, i.e., a voxel has $2 \times D$ neighbors (e.g., 4 for 2D and 6 for 3D images) or {\em all} connectivity, i.e., a voxel has $3^D-1$ neighbors (e.g., 8 for 2D and 26 for 3D images). In 2D, direct connectivity connects pixels that share an edge, and all connectivity additionally connects diagonally adjacent pixels, with their boundary only sharing a vertex. For 3D data, direct connectivity connects pixels that share a surface cell, and all-connectivity also connects pixels whose boundaries only intersect in a line or a vertex on the boundary. To ensure that the Jordan-Curve Theorem holds, it is necessary to use opposite connectivity choices for foreground and background, e.g., all connectivity for the foreground and direct connectivity for the background. In this work, we use the letter \textbf{A} to denote the setting where all connectivity is applied to the foreground and direct connectivity is applied to the background. We denote the inverted connectivity choice with the letter \textbf{D}.

In cubical complexes, which are often used to describe the topology of digital images \cite{clough2020topological,stucki2023topologically,lux2024topograph,hu2019topology}, the complex's construction choice implicitly encodes a specific connectivity. V-construction $V(\mathcal{I})$ is closely related to direct connectivity for the foreground and all connectivity for the background. The T-construction $T(\mathcal{I})$ is closely related to the inverted connectivity choice. Bleile et al. \cite{bleile2022persistent} describe the relationship between the two constructions. 

Making the correct connectivity choice is essential to accurately capture the domain-specific semantics of the underlying data. Investigating the semantics of the most commonly used datasets, we find that often one connectivity choice is favorable over the other. For example, in the frequently used DRIVE dataset, the smallest vessels are represented through pixels that are only linked through all connectivity (see Figure \ref{fig:2_1_connectivity} (a)). There, choosing direct connectivity instead of all connectivity results in a more than 100-fold increase in the number of foreground connected components (see Table \ref{tab:CC_tab}) because single vessel segments are split into multiple connected components. Among the most important previous works, only one paper reports their connectivity choice, which is the semantically unfavorable \textbf{D}-connectivity for the DRIVE dataset; most other works do not explicitly state their connectivity choices \cite{attari2023multi,wu2024deep,stucki2023topologically,hu2021topology,hu2019topology,li2023robust,li2024universal,berger2024topologically,hu2022structure}. This is problematic for the commonly used datasets, as their topological representation strongly depends on the connectivity choices, as displayed in Table \ref{tab:CC_tab}. However, not all datasets exhibit this problem, e.g., MSSEG2 \cite{commowick2021msseg}.

\textbf{Experimental investigation of the problem:} To demonstrate the effect of the connectivity choice on topology-aware segmentation methods, we perform an experiment in which we train and evaluate different methods with the two distinct connectivity choices \textbf{A} and \textbf{D}. The results are displayed in Table \ref{tab:con_exp}. 
We observe that the variation originating from connectivity choices dominates the inter-method variation. The ranking of the methods changes drastically depending on the connectivity choice. We conclude that making the semantically meaningful connectivity choice is essential for a robust evaluation. High performance under the unfavorable \textbf{A} connectivity does not indicate that high performance can be expected under the more meaningful \textbf{D} connectivity. Interestingly, we mostly observe negative correlations for the scores under the distinct connectivity choices. A potential explanation is that methods that are independent of the connectivity choice during training (Dice, Mosin, clDice) perform very well compared to methods affected by the connectivity choice (HuTopo, BettiM, TopoG) even in the semantically unfavorable setting. However, when the favorable connectivity choice is made, the latter methods capture the semantically expected topology and, therefore, achieve higher performance.


\begin{table}[t]
    \centering
    \scriptsize
    \caption{Results of the connectivity experiment on the CREMI dataset. The first block of results stems from evaluation and training with direct connectivity for the foreground and all connectivity for the background (D). The next block contains the results for the inverted setting (A). 
    Numbers in brackets () behind the results contain the method's ranking. The bottom block contains an analysis of the comparability of a method's rank under the two connectivity choices.}
    \label{tab:con_exp}
    \begin{tabular}{llccccccc}
    \toprule
     & \textbf{Loss} & \textbf{DICE $\uparrow$} & \textbf{B0 $\downarrow$} & \textbf{B1 $\downarrow$} & \textbf{BM0 $\downarrow$} & \textbf{BM1 $\downarrow$} & \textbf{VOI $\downarrow$} & \textbf{ARE $\downarrow$} \\ 
    \midrule
    \multirow{5}{*}{\textbf{D} \cmark} 
        & Dice       & .9466 \textcolor{c1}{\textbf{(1)}} & 1.52 \textcolor{c6}{\textbf{(6)}} & 2.09 \textcolor{c2}{\textbf{(2)}} & 2.38 \textcolor{c6}{\textbf{(6)}} & 4.14 \textcolor{c1}{\textbf{(1)}} & .4212 \textcolor{c1}{\textbf{(1)}} & .1129 \textcolor{c1}{\textbf{(1)}} \\
        & ClDice     & .9404 \textcolor{c5}{\textbf{(6)}} & 1.45 \textcolor{c5}{\textbf{(5)}} & 2.44 \textcolor{c4}{\textbf{(4)}} & 2.32 \textcolor{c5}{\textbf{(5)}} & 4.31 \textcolor{c4}{\textbf{(4)}} & .4580 \textcolor{c5}{\textbf{(5)}} & .1271 \textcolor{c5}{\textbf{(5)}} \\
        & HuTopo     & .9455 \textcolor{c2}{\textbf{(2)}} & 1.34 \textcolor{c4}{\textbf{(4)}} & 2.11 \textcolor{c3}{\textbf{(3)}} & 2.08 \textcolor{c4}{\textbf{(4)}} & 4.24 \textcolor{c2}{\textbf{(2)}} & .4270 \textcolor{c2}{\textbf{(2)}} & .1129 \textcolor{c2}{\textbf{(2)}} \\
        & BettiM     & .9438 \textcolor{c4}{\textbf{(4)}} & 1.21 \textcolor{c1}{\textbf{(1)}} & 2.06 \textcolor{c1}{\textbf{(1)}} & 1.95 \textcolor{c2}{\textbf{(2)}} & 4.30 \textcolor{c3}{\textbf{(3)}} & .4336 \textcolor{c4}{\textbf{(4)}} & .1140 \textcolor{c3}{\textbf{(3)}} \\
        & Mosin      & .9435 \textcolor{c6}{\textbf{(5)}} & 1.30 \textcolor{c3}{\textbf{(3)}} & 2.74 \textcolor{c6}{\textbf{(6)}} & 1.90 \textcolor{c1}{\textbf{(1)}} & 4.59 \textcolor{c6}{\textbf{(6)}} & .4881 \textcolor{c6}{\textbf{(6)}} & .1370 \textcolor{c6}{\textbf{(6)}} \\
        & TopoG      & .9444 \textcolor{c3}{\textbf{(3)}} & 1.26 \textcolor{c2}{\textbf{(2)}} & 2.46 \textcolor{c5}{\textbf{(5)}} & 2.01 \textcolor{c3}{\textbf{(3)}} & 4.33 \textcolor{c5}{\textbf{(5)}} & .4324 \textcolor{c3}{\textbf{(3)}} & .1183 \textcolor{c4}{\textbf{(4)}} \\ 
    \midrule
    \multirow{5}{*}{\textbf{A} \xmark}
        & Dice       & .9435 \textcolor{c3}{\textbf{(3)}} & 0.46 \textcolor{c2}{\textbf{(2)}} & 3.91 \textcolor{c4}{\textbf{(4)}} & 0.51 \textcolor{c1}{\textbf{(1)}} & 7.39 \textcolor{c3}{\textbf{(3)}} & .4397 \textcolor{c3}{\textbf{(3)}} & .1203 \textcolor{c3}{\textbf{(3)}} \\
        & ClDice     & .9441 \textcolor{c2}{\textbf{(2)}} & 0.53 \textcolor{c4}{\textbf{(4)}} & 3.57 \textcolor{c1}{\textbf{(1)}} & 0.58 \textcolor{c3}{\textbf{(3)}} & 7.20 \textcolor{c1}{\textbf{(1)}} & .4336 \textcolor{c2}{\textbf{(2)}} & .1169 \textcolor{c2}{\textbf{(2)}} \\
        & HuTopo     & .9418 \textcolor{c4}{\textbf{(4)}} & 0.46 \textcolor{c1}{\textbf{(1)}} & 4.36 \textcolor{c6}{\textbf{(6)}} & 0.54 \textcolor{c2}{\textbf{(2)}} & 7.66 \textcolor{c5}{\textbf{(5)}} & .4481 \textcolor{c4}{\textbf{(4)}} & .1247 \textcolor{c6}{\textbf{(6)}} \\
        & BettiM     & .9411 \textcolor{c6}{\textbf{(6)}} & 0.57 \textcolor{c6}{\textbf{(6)}} & 3.96 \textcolor{c5}{\textbf{(5)}} & 0.63 \textcolor{c6}{\textbf{(6)}} & 7.81 \textcolor{c6}{\textbf{(6)}} & .4547 \textcolor{c6}{\textbf{(6)}} & .1219 \textcolor{c4}{\textbf{(4)}} \\
        & Mosin      & .9450 \textcolor{c1}{\textbf{(1)}} & 0.53 \textcolor{c3}{\textbf{(3)}} & 3.66 \textcolor{c2}{\textbf{(2)}} & 0.61 \textcolor{c4}{\textbf{(4)}} & 7.23 \textcolor{c2}{\textbf{(2)}} & .4282 \textcolor{c1}{\textbf{(1)}} & .1141 \textcolor{c1}{\textbf{(1)}} \\
        & TopoG      & .9417 \textcolor{c5}{\textbf{(5)}} & 0.56 \textcolor{c5}{\textbf{(5)}} & 3.86 \textcolor{c3}{\textbf{(3)}} & 0.62 \textcolor{c5}{\textbf{(5)}} & 7.50 \textcolor{c4}{\textbf{(4)}} & .4489 \textcolor{c5}{\textbf{(5)}} & .1242 \textcolor{c5}{\textbf{(5)}} \\ 
    \midrule \midrule
    \multicolumn{9}{c}{\textbf{Comparison of Method Ranking for D and A Connectivity Choices}} \\ \midrule
      \multicolumn{2}{l}{\textbf{Spearman's $\rho$}}    & \textbf{-0.37}    & \textbf{-0.85}    & \textbf{-0.66}     & \textbf{-0.77}   & \textbf{-0.37}    & \textbf{-0.43}   &  \textbf{-0.70} \\ 
    \multicolumn{2}{l}{\textbf{Kendall's $\tau$}}       & \textbf{-0.07}    & \textbf{-0.79}    & \textbf{-0.47}     & \textbf{-0.60}   & \textbf{-0.20}    & \textbf{-0.20}   &  \textbf{-0.55} \\ 
    \multicolumn{2}{l}{\textbf{Pearsons's $r$}}           & \textbf{-0.34}    & \textbf{-0.70}    & \textbf{-0.69}     & \textbf{-0.76}   & \textbf{-0.39}    & \textbf{-0.77}   &  \textbf{-0.87} \\ 
    \bottomrule
    \end{tabular}
\end{table}

\textbf{Strategies to identify and resolve connectivity problems:} We encourage future works to make individual and semantics-oriented connectivity choices for every individual dataset. For the three common benchmarking datasets, we determine \textbf{A} connectivity for DRIVE (to maintain vessel connectivity), \textbf{D} connectivity for CREMI (to avoid separation of synaptic clefts), and \textbf{D} connectivity for Roads (to minimize connectivity artifacts in the background, see Section \ref{sec:artifacts}) as the sensible connectivity choices. The choices should be made transparent to the reader when reporting metrics; e.g., for Betti numbers $\beta_{0_A}$ versus $\beta_{0_D}$; alternatively $\beta_{0_T}$ versus $\beta_{0_V}$ for an explicit notation of the cubical complex construction. This notation can be extended to all metrics where connectivity choices are influential, e.g., $ARE_A$ or $ARE_D$. 
Moreover, we propose a simple method to see how susceptible a dataset is to connectivity choices. We consider the two different partitions 
\begin{align*}
P_{D} = \{F_1,F_2, ..., F_k, B_1,B_2, ..., B_l\}\\
P_{A} = \{F'_1,F'_2, ..., F'_n, B'_1, B'_2, ..., B'_m\}
\end{align*}
of a label $G$ that result from connected component labeling with the \textbf{D} and \textbf{A} connectivity. The two partitions are the basis for calculating metrics (e.g., $\beta$, ARE, VOI). We define the connectivity susceptibility for a metric, e.g., $\beta_0^{err}$, as 
\begin{equation}
    \beta^{err}_{0_{D \text{vs.}A}} = |\beta_0(P_{D}) - \beta_0(P_{A})|
\end{equation}
where $\beta_0(P)$ measures the number of foreground components in $P$. Note that because $\beta_0^{err}$ is a \textit{true} metric, the results are independent of the order of $P_{D}$ and $P_{A}$. Other susceptibility metrics are defined accordingly, and large values indicate high susceptibility. Table \ref{tab:suscept_tab} shows the results of this analysis for different datasets and metrics. The results of the connectivity experiment (see Table \ref{fig:2_1_connectivity}) show that high susceptibility values are indicators for a large discrepancy of scores under \textbf{D} and \textbf{A} connectivity.

\begin{table}[t]
    \centering
    \scriptsize
    \renewcommand{\arraystretch}{1.1}
    \setlength{\tabcolsep}{7pt}
    \caption{Summary of the susceptibility to connectivity choices, introduced by, e.g., using the T- or V-construction, for different datasets. High scores indicate that a dataset, combined with a specific metric, is highly susceptible to connectivity choice.}
    \label{tab:suscept_tab}
    \begin{tabular}{lcccc}
        \toprule
        \textbf{Dataset} & $\mathbf{\beta^{err}_{0_{D \text{vs.}A}}}$ & $\mathbf{\beta^{err}_{1_{D \text{vs.}A}}}$ & $\mathbf{VOI_{D \text{vs.}A}}$ & $\mathbf{ARE_{D \text{vs.}A}}$  \\
        \midrule \midrule
        \textbf{\quad DRIVE} &  467.95   & 32.23  & 0.1171 & 0.8608\\
        \textbf{\quad CREMI} &  0.056   & 38.216 & 0.0025 & 0.0004 \\
        \textbf{\quad Roads} &  0.0242   & 10.5161 & 0.0030 & 0.0009  \\
        \bottomrule
        \vspace{-\intextsep}
    \end{tabular}

\end{table}

\subsection{\textit{Topological Artifacts} in the label skew evaluation results}
\label{sec:artifacts} 

\begin{figure}[t]
    \centering
    \includegraphics[width=0.8\linewidth, trim={0 0.8cm 0 0.85cm},clip]{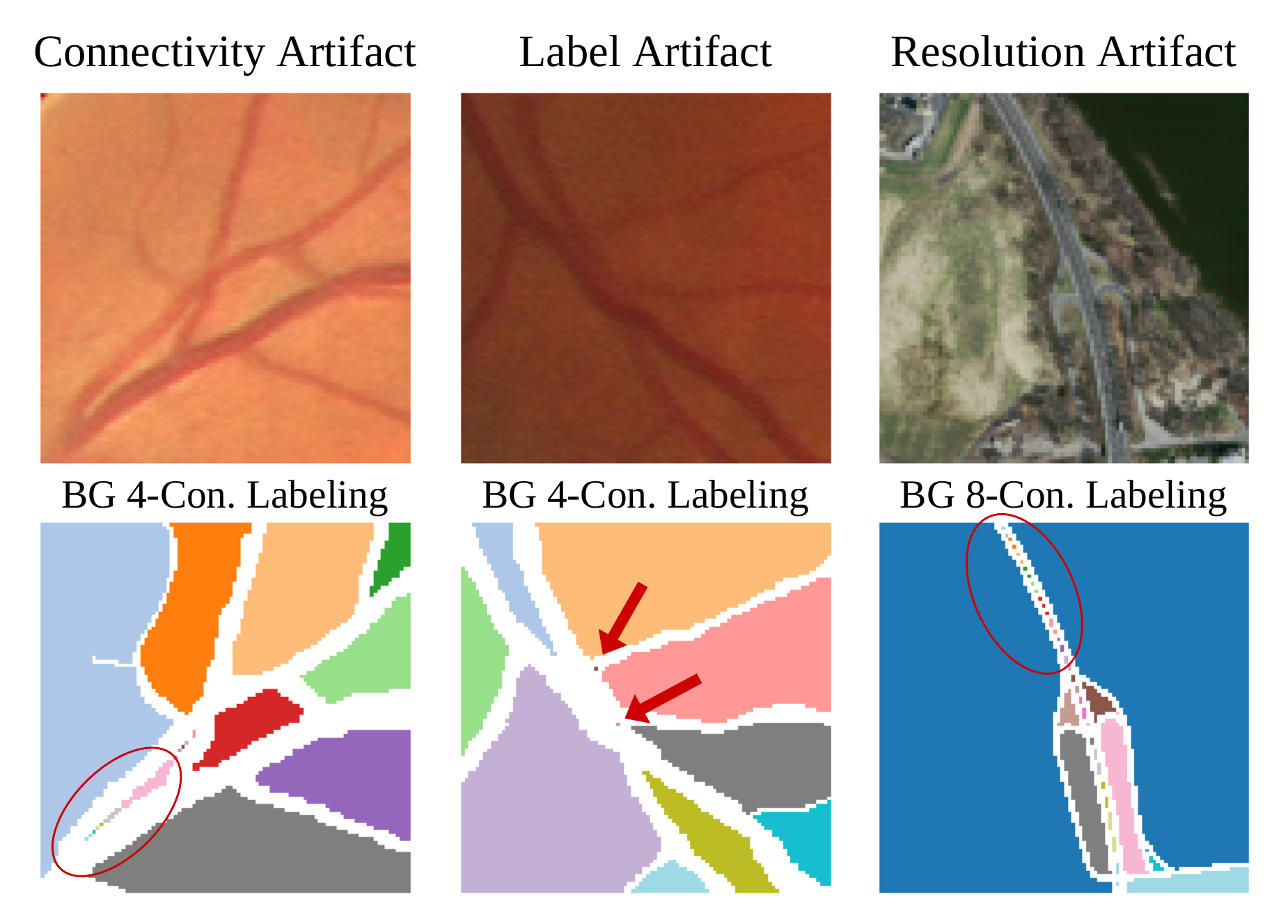}
    \caption{Examples of topological artifacts. The two left columns show connectivity artifacts and label noise in the DRIVE dataset. The right column shows resolution artifacts in the Roads dataset. Arrows and circles indicate topological artifacts.}
    \label{fig:2_2_artifacts}
\end{figure}

In the context of topology-aware image segmentation, we propose to denote \textit{topological artifacts} as topological features existing in the ground truth labels but \textit{conveying no or a wrong semantic meaning}. We observe three prominent causes of such artifacts in the investigated datasets: (1) connectivity issues, (2) label noise, and (3) insufficient resolution (see Fig. \ref{fig:2_2_artifacts}).

\textit{Connectivity artifacts:} In many datasets, neither \textbf{A} nor \textbf{D} connectivity resolve all connectivity issues. In these cases, we define connectivity artifacts as topological artifacts that occur due to the choices made in connectivity. Connectivity artifacts can be found in the DRIVE dataset (see Figure \ref{fig:2_2_artifacts}). While the foreground 8-connectivity is absolutely crucial to capture the semantics of the small vessels for the DRIVE dataset, the then required background 4-connectivity divides a single inter-vessel area into numerous separate components. In the DRIVE dataset, these artifacts result in almost a doubling of background components for 4-connectivity instead of 8-connectivity (see Table \ref{tab:art_count}). 

\textit{Label noise artifacts:} Label noise artifacts occur due to annotation errors during the generation of the ground truth segmentation mask. While some types of label noise do not interfere with the topological representation, other annotation errors can lead to topological artifacts. Single-pixel label noise is a common phenomenon in segmentation labels. These errors only have a negligible impact on pixel-wise metrics. However, the impact on the topological representation can be dramatic. 

\textit{Resolution artifacts:} We describe resolution artifacts as artifacts caused by the representation of the underlying semantics with insufficient resolution. Connectivity artifacts are, in fact, a special case of the insufficient resolution issue. An example of resolution artifacts can be found in the Roads dataset, where the insufficient resolution of the aerial images causes the separating strip of two lanes to appear as dozens of separated background components, as displayed in Figure \ref{fig:2_2_artifacts}. These artifacts often have negligible effects on pixel-wise accuracy but can drastically affect the evaluation with common topological metrics and interfere with the optimization mechanisms of topology-aware methods. 

\textbf{Experimental investigation of the problem:} We compare the performance of different topology-aware methods before and after removing small background components using the DRIVE dataset. Visual inspection revealed that these components are predominantly caused by topological label noise and connectivity artifacts. The results of this experiment indicate a strong effect on the Betti and Betti matching metrics (see Table \ref{tab:artifact_exp}). Particularly for Betti number 1 error ($\sim$-29\%) and Betti matching 1 error ($\sim$-43\%), we see extreme changes caused by the artifacts (see Figure \ref{fig:2_2_artifacts} for examples). In comparison, the artifacts only induce minute changes for the Dice score and VOI and ARE, which have traits of both topological and pixel-wise metrics.


\begin{table}[t]
    \scriptsize
    \centering
    \caption{Results of the artifacts experiment on DRIVE. The first results block stems from evaluation and training with adapted ground truth where small components ($\leq5$ pixel), mostly resembling topological artifacts, are removed. The next block contains the results for an adapted ground truth where small components ($\leq5$ pixel), mostly resembling topological artifacts, are removed. 
    Numbers in brackets () contain the method's ranking. The bottom block contains an analysis of the comparability of a method's rank before and after removing small components.}
    \label{tab:artifact_exp}
    \begin{tabular}{llccccccc}
    \toprule
     & \textbf{Loss} & \textbf{DICE $\uparrow$} & \textbf{B0$_A$ $\downarrow$} & \textbf{B1$_A$  $\downarrow$} & \textbf{BM0$_A$  $\downarrow$} & \textbf{BM1$_A$  $\downarrow$} & \textbf{VOI$_A$  $\downarrow$} & \textbf{ARE$_A$  $\downarrow$} \\ 
    \midrule
    \multirow{5}{*}{\begin{tabular}[c]{@{}l@{}}\textbf{Corrected}\\ \textbf{Label}\end{tabular} \cmark}
        & Dice       & .8808 \textcolor{c4}{\textbf{(4)}} & 9.05 \textcolor{c5}{\textbf{(5)}} & 6.35 \textcolor{c4}{\textbf{(4)}} & 9.59   \textcolor{c6}{\textbf{(6)}} & 11.88 \textcolor{c2}{\textbf{(2)}} & .5187 \textcolor{c6}{\textbf{(6)}} & .2550 \textcolor{c6}{\textbf{(6)}} \\
        & ClDice     & .8875 \textcolor{c1}{\textbf{(1)}} & 7.29 \textcolor{c4}{\textbf{(4)}} & 6.61 \textcolor{c6}{\textbf{(6)}} & 7.81   \textcolor{c4}{\textbf{(4)}} & 11.75 \textcolor{c1}{\textbf{(1)}} & .4869 \textcolor{c1}{\textbf{(1)}} & .2277 \textcolor{c1}{\textbf{(1)}} \\
        & HuTopo     & .8816 \textcolor{c3}{\textbf{(3)}} & 2.65 \textcolor{c1}{\textbf{(1)}} & 4.50 \textcolor{c2}{\textbf{(2)}} & 3.12   \textcolor{c1}{\textbf{(1)}} & 13.72 \textcolor{c6}{\textbf{(6)}} & .5022 \textcolor{c2}{\textbf{(2)}} & .2473 \textcolor{c4}{\textbf{(4)}} \\
        & BettiM     & .8786 \textcolor{c6}{\textbf{(6)}} & 2.70 \textcolor{c2}{\textbf{(2)}} & 4.34 \textcolor{c1}{\textbf{(1)}} & 3.17   \textcolor{c2}{\textbf{(2)}} & 12.25 \textcolor{c5}{\textbf{(5)}} & .5129 \textcolor{c5}{\textbf{(5)}} & .2518 \textcolor{c5}{\textbf{(5)}} \\
        & Mosin      & .8823 \textcolor{c2}{\textbf{(2)}} & 9.39 \textcolor{c6}{\textbf{(6)}} & 6.47 \textcolor{c5}{\textbf{(5)}} & 9.39   \textcolor{c5}{\textbf{(5)}} & 12.21 \textcolor{c4}{\textbf{(4)}} & .5069 \textcolor{c3}{\textbf{(3)}} & .2380 \textcolor{c2}{\textbf{(2)}} \\
        & TopoG      & .8804 \textcolor{c5}{\textbf{(5)}} & 4.68 \textcolor{c3}{\textbf{(3)}} & 6.11 \textcolor{c3}{\textbf{(3)}} & 5.08   \textcolor{c3}{\textbf{(3)}} & 11.91 \textcolor{c3}{\textbf{(3)}} & .5078 \textcolor{c4}{\textbf{(4)}} & .2413 \textcolor{c3}{\textbf{(3)}} \\ 
    \midrule
    \multirow{5}{*}{\begin{tabular}[c]{@{}l@{}}\textbf{Original}\\ \textbf{Label}\end{tabular} \xmark} 
        & Dice       & .8873 \textcolor{c1}{\textbf{(1)}} & 7.92 \textcolor{c4}{\textbf{(4)}} & 11.02 \textcolor{c4}{\textbf{(4)}} & 8.44  \textcolor{c4}{\textbf{(4)}} & 16.38 \textcolor{c2}{\textbf{(2)}} & .4910 \textcolor{c2}{\textbf{(2)}} & .2336 \textcolor{c2}{\textbf{(2)}} \\
        & ClDice     & .8827 \textcolor{c4}{\textbf{(4)}} & 9.76 \textcolor{c6}{\textbf{(6)}} & 11.47 \textcolor{c6}{\textbf{(6)}} & 10.31 \textcolor{c6}{\textbf{(6)}} & 16.67 \textcolor{c4}{\textbf{(4)}} & .5026 \textcolor{c4}{\textbf{(4)}} & .2375 \textcolor{c4}{\textbf{(4)}} \\
        & HuTopo     & .8827 \textcolor{c5}{\textbf{(5)}} & 4.33 \textcolor{c2}{\textbf{(2)}} & 8.30  \textcolor{c2}{\textbf{(2)}} & 4.87  \textcolor{c2}{\textbf{(2)}} & 18.78 \textcolor{c6}{\textbf{(6)}} & .4984 \textcolor{c3}{\textbf{(3)}} & .2375 \textcolor{c3}{\textbf{(3)}} \\
        & BettiM     & .8831 \textcolor{c2}{\textbf{(2)}} & 2.44 \textcolor{c1}{\textbf{(1)}} & 7.56  \textcolor{c1}{\textbf{(1)}} & 2.91  \textcolor{c1}{\textbf{(1)}} & 18.16 \textcolor{c5}{\textbf{(5)}} & .4894 \textcolor{c1}{\textbf{(1)}} & .2277 \textcolor{c1}{\textbf{(1)}} \\
        & Mosin      & .8828 \textcolor{c3}{\textbf{(3)}} & 8.35 \textcolor{c5}{\textbf{(5)}} & 11.39 \textcolor{c5}{\textbf{(5)}} & 8.88  \textcolor{c5}{\textbf{(5)}} & 16.57 \textcolor{c3}{\textbf{(3)}} & .5056 \textcolor{c5}{\textbf{(5)}} & .2384 \textcolor{c5}{\textbf{(5)}} \\
        & TopoG      & .8746 \textcolor{c6}{\textbf{(6)}} & 5.11 \textcolor{c3}{\textbf{(3)}} & 10.38 \textcolor{c3}{\textbf{(3)}} & 5.65  \textcolor{c3}{\textbf{(3)}} & 16.36 \textcolor{c1}{\textbf{(1)}} & .5355 \textcolor{c6}{\textbf{(6)}} & .2658 \textcolor{c6}{\textbf{(6)}} \\ 
    \midrule \midrule
    \multicolumn{9}{c}{\textbf{Comparison of Performance Label \xmark \ and Fixed Label \cmark}} \\ \midrule
    \multicolumn{2}{l}{\textbf{Avg. Difference}}         & \textbf{.0003}    & \textbf{-0.36}      & \textbf{-4.29}       & \textbf{-0.48}     & \textbf{-4.87}       & \textbf{.0022}     &  \textbf{.0034} \\ 
    \multicolumn{2}{l}{\textbf{Avg. Rel. Change }}       & \textbf{0.38\%}    & \textbf{-5.86\%}    & \textbf{-42.91\%}     & \textbf{-6.99\%}   & \textbf{-28.33\%}    & \textbf{0.53\%}     &  \textbf{1.72\%} \\ 
    \bottomrule
    \end{tabular}
\end{table}

\textbf{Strategies to mitigate the problem:} A visual inspection of the dataset is paramount to identify topological artifacts and identify ways to remove them. While a visual inspection of the DRIVE dataset reveals that small connected components are mostly artifacts, in the CREMI dataset, such components often reflect the beginning of neurons or synaptic clefts, and removing them would destroy essential topological information. For the DRIVE and Roads dataset, a simple image processing that removes those isolated pixels is an effective remedy and allows for a focus on the semantically meaningful topologic structures. 

\begin{table}[b]
\centering
\scriptsize
\renewcommand{\arraystretch}{1.1}
\setlength{\tabcolsep}{7pt}
\caption{Effect of removal of components up to a specific size on the topological representation of the commonly used CREMI, roads, and DRIVE dataset. We use the semantically favorable connectivities for each dataset; Drive: \textbf{A}, CREMI and Roads: \textbf{D}}
\label{tab:art_count}
\begin{tabular}{lcccccccc}
\toprule
 & \multicolumn{2}{c}{\textbf{DRIVE$_A$}} & \multicolumn{2}{c}{\textbf{Roads$_D$}} & \multicolumn{2}{c}{\textbf{CREMI$_D$}} \\
\cmidrule(lr){2-3} \cmidrule(lr){4-5} \cmidrule(lr){6-7}
\textbf{Components} & \textbf{FG} & \textbf{BG} & \textbf{FG} & \textbf{BG} & \textbf{FG} & \textbf{BG} \\
\midrule
No Removal        & 132 & 2362 & 644 & 7535 & 712 & 44815 \\
1 Pix. Removal    & 125 & 1839  & 641 & 7124 & 712 & 42826 \\
2 Pix. Removal    & 120 & 1706  & 635 & 6905 & 712 & 42134 \\
5 Pix. Removal    & 117 & 1629  & 613 & 6664 & 712 & 40795 \\
Min/Max Ratio  & 88.6\% & 69.0\% & 95.0\%  & 88.4\% & 100\% & 91.0\%   \\
\bottomrule
\end{tabular}

\end{table}

\subsection{\textit{Evaluation Metrics}, as commonly reported, lack expressive power}

\begin{figure}[t]
    \centering
    \includegraphics[width=0.9\linewidth, trim={0 0.7cm 0 0.85cm},clip]{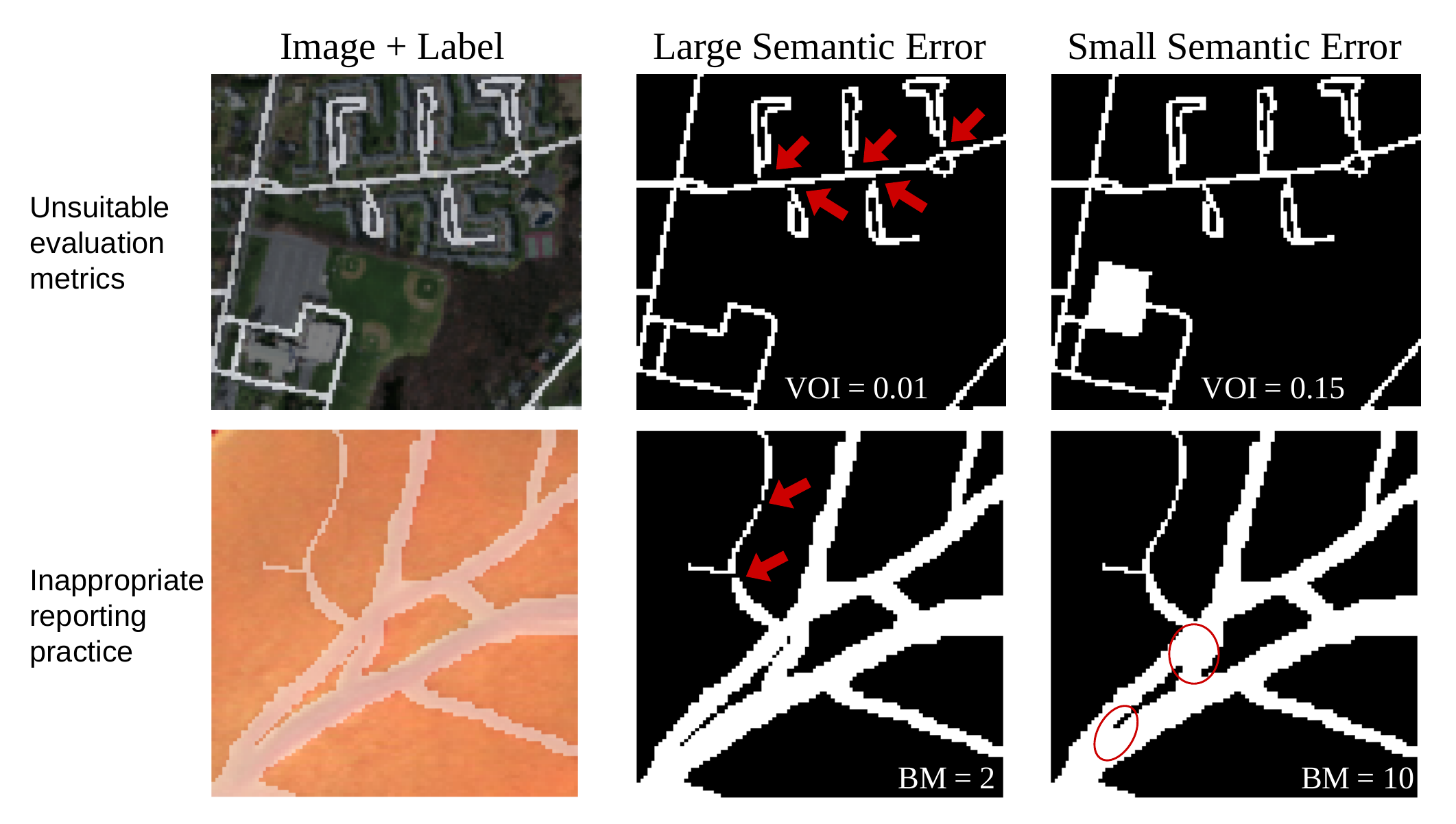}
    \caption{Examples for the impact of \textit{inappropriate reporting practice} on DRIVE (top) and \textit{unsuitable evaluation metrics} on the Roads dataset (bottom). Left: Image with an overlay of the segmentation label. Middle: Unfavorable predictions, with detached vessels $BM =2$ (top) and disconnected residential blocks $VOI = 0.01$(bottom). Right: Favorable predictions with missing background components without semantic meaning $BM =10$ (top) and additional segmentation of parking areas $VOI =0.15$ (bottom).} 
    \label{fig:2_3_metrics}
\end{figure}

The evaluation of topology-aware segmentation typically employs distributional (e.g., VOI or ARE/ARI) and/or topological metrics, such as the Betti number or Betti matching error in combination with pixel-wise/overlap-based metrics (e.g., Dice or cross-entropy). We find that current reporting practices often fail to provide an expressive and interpretable characterization of topological correctness.

\textbf{Experimental investigation of the problem:} Distributional metrics such as VOI and ARI/ARE irreversibly entangle topological and volumetric errors \cite{funke2017ted,nunez2013machine}. Therefore, reporting only these metrics in addition to pixel-wise performance metrics does not allow for an expressive evaluation of topological accuracy. Figure \ref{fig:2_3_metrics} shows an example where volumetric deviations largely dominate the score of these metrics while small yet critical topological errors are marginalized. Empirically, we validate this shortcoming and find that the distributional metrics do not always correlate with topological accuracy. For example, we find no correlation between BM0$_A$ and ARE$_A$ ($\rho=-0.03$) or VOI$_A$ ($\rho=0.31$) in the DRIVE dataset. Here, BM0$_A$ is the most important topological metric because it captures disconnected or incorrect vessel segments. However, in other cases, we find a positive correlation. For CREMI, BM1$_D$ is an important topological metric since it captures false splits and merges of neurons. Here, BM1$_D$ correlates well with ARE$_D$ (Spearman's $\rho=0.94$) and VOI$_D$ ($\rho=0.83$).

Betti number errors are a better alternative to distributional metrics as they disentangle any volumetric effects and provide topologically interpretable values. However, Stucki et al. \cite{stucki2023topologically} show that merely comparing the Betti numbers can be misleading as it disregards the spatial correspondence of invariants and propose the Betti matching error. We investigate this statement empirically and mostly find a good correspondence of performance rankings between Betti number and Betti matching errors, e.g. in CREMI (B0$_D$ with BM0$_D$ ($\rho=0.83$), and B1$_D$ with BM1$_D$ ($\rho=0.94$)). However, we also observe a negative correlation between B1$_A$ and BM1$_A$ ($\rho=-0.77$) for the DRIVE dataset. Here, the Betti number 1 error of some methods is reduced by features without spatial correspondence. 

While the Betti number and Betti matching error disentangle volumetric effects and provide interpretable values, they are often aggregated across dimensions (e.g., $\beta^{err} = \beta^{err}_0 + \beta^{err}_1$) \cite{hu2021topology,attari2023multi,hu2019topology,hu2022structure,berger2024topologically,stucki2023topologically,lux2024topograph}. Figure \ref{fig:2_3_metrics} illustrates how the combined Betti matching error can be misleading. The segmentation to the right has a 5-times higher Betti number error but is preferable for downstream network analysis tasks because it maintains vessel connectivity. In our empirical analyses, we find that BM0$_D$ and BM1$_D$ show no positive rank correlation ($\rho=-0.77$, see Table \ref{tab:con_exp}). Here, BM1$_D$ is the most important topological metric as it shows almost perfect rank correlation with ARI$_D$ ($\rho=0.94$) and VOI$_D$ ($\rho=0.83$), which were found to correlate well with neuron segmentation quality by domain experts \cite{arganda2015crowdsourcing}. Therefore, aggregating BM0$_D$ and BM1$_D$ reduces expressive power.

\textbf{Strategies to mitigate the problem:} We propose three important reporting practices for evaluating topology-aware image segmentation models. (1) We recommend always reporting at least one disentangled pair of purely topological and volumetric metrics. Due to the occasional issues with spatial correspondence, we propose to use Betti matching errors instead of Betti number errors (e.g., BM0, BM1, and Dice). (2) We recommend reporting topological errors always without aggregation across dimensions (e.g., BM0, BM1, instead of BM). (3) Finally, we recommend a problem-aware selection of additional metrics \cite{maier2024metrics} (e.g., ARI and VOI for Neuron Segmentation and clDice for Vessel Segmentation).

\section{Discussion}
The presented work sheds light on common pitfalls during the evaluation of topology-aware image segmentation methods. We identify (1) connectivity definition, (2) label artifacts, and (3) misaligned use of evaluation metrics as major problems that heavily impact previous studies. In dedicated experiments, we show that these pitfalls are a major limitation for meaningful benchmarking in topology-aware image segmentation. Specifically, our results indicate that topological performance rankings are vulnerable to connectivity choices, showing on average negative correlations of the rankings (avg. Spearman's $\rho$ = -0.63) of \textbf{A} and \textbf{D} connectivity. We find that label artifacts can comprise up to 43\% of measured topological errors in some metrics. Finally, we find that flawed evaluation practices, such as an aggregation of Betti numbers, drastically impair the expressivity of the evaluation. 

Based on our analysis, we conclude with the following recommendations for future works: (1) \textit{Connectivity choices must be made on a dataset and not on a method basis. The choices have to be transparent for the reader}. We introduce a new method to quantify the connectivity susceptibility of datasets, providing a measure of the importance of connectivity choices. (2) \textit{Topological artifacts must be considered in topology-aware image segmentation}. We provide a definition for topological artifacts that were previously overlooked due to their negligible influence on pixel-wise evaluation. (3) \textit{Evaluation metrics should disentangle volumetric and topological information and topological errors of different dimensions. Other metrics should be added in a problem-oriented manner.} Ultimately, this paper should ignite a discussion on the current state and best practices of topology-aware image segmentation.

\paragraph{Limitations.} While the presented issues are valid for 2D as well as 3D images, an empirical evaluation of 3D datasets is still warranted because the commonly used slicing of 3D volumes in 2D images changes a segmentation's topological requirements. Furthermore, our work deliberately does not discuss general pitfalls in model evaluation that are not specific to topology-aware image segmentation. These general pitfalls include insufficient variation analysis \cite{christodoulou2024confidence}, unfair baseline comparisons \cite{isensee2024nnu}, or general unsuitability of metrics \cite{reinke2024understanding}.

\begin{credits}
\subsubsection{\ackname} AB is supported by the Stiftung der Deutschen Wirtschaft. MM is funded by the German Research Foundation under project 532139938.

\subsubsection{\discintname}
The authors have no competing interests to declare that are
relevant to the content of this article.
\end{credits}

%
%
\bibliographystyle{splncs04}
\bibliography{references.bib}

\end{document}